\documentclass[10pt,twocolumn,letterpaper]{article}

\usepackage{cvpr}              

\usepackage{graphicx}
\usepackage{amsmath}
\usepackage{amssymb}
\usepackage{booktabs}
\usepackage[ruled,linesnumbered]{algorithm2e}

\usepackage[pagebackref,breaklinks,colorlinks]{hyperref}

\usepackage[capitalize]{cleveref}
\crefname{section}{Sec.}{Secs.}
\Crefname{section}{Section}{Sections}
\Crefname{table}{Table}{Tables}
\crefname{table}{Tab.}{Tabs.}

\newlength\mylen
\let\oldnl\nl
\newcommand{\nonl}{\renewcommand{\nl}{\let\nl\oldnl}}
\newcommand\myinput[1]{%
  \settowidth\mylen{\KwIn{}}%
  \setlength\hangindent{\mylen}%
  \nonl\hspace*{\mylen}#1\\}

\def\cvprPaperID{5} 
\def\confName{CVPR}
\def\confYear{2022}

\begin{document}

\title{SMM-Conv: Scalar Matrix Multiplication with Zero Packing for\\ Accelerated Convolution}


\author{Amir Ofir\\
Ariel University\\
Ariel, Israel\\
{\tt\small amir.ofir@msmail.ariel.ac.il}
\and
Gil Ben-Artzi\\
Ariel University\\
Ariel, Israel\\
{\tt\small gilba@g.ariel.ac.il}
}
\maketitle

\begin{abstract}
We present a novel approach for accelerating convolutions during inference for CPU-based architectures. The most common method of computation involves packing the image into the columns of a matrix (im2col) and performing general matrix multiplication (GEMM) with a matrix of weights. This results in two main drawbacks: (a) im2col requires a large memory buffer and can experience inefficient memory access, and (b) while GEMM is highly optimized for scientific matrices multiplications, it is not well suited for convolutions. We propose an approach that takes advantage of scalar-matrix multiplication and reduces memory overhead. Our experiments with commonly used network architectures demonstrate a significant speedup compared to existing indirect methods. 
\end{abstract}

\section{Introduction}
\label{sec:intro}

A major limitation of Convolutional Neural Networks (CNN) on mobile and low-power devices is the high computational cost associated with chains of convolutional layers~\cite{vanhouckeimproving,krizhevsky2012imagenet,10095537}. As a result, their availability for many image-related tasks, such as denoising~\cite{9380520}, can not be extended to many common consumer devices since they are not equipped with high-end GPUs.

Convolutional layers can be computed~\cite{jia2014caffe, pytorch_9015} using general matrix multiplication (\emph{GEMM}), a matrix multiplication procedure found in the majority of computational libraries \cite{blackford2002updated}. Matrix multiplication based convolutions are popular as the GEMM is heavily optimized by CPU vendors. It exploits CPU caching and register manipulation for continuous computation and fused multiplication accumulation (\emph{FMA}) for conducting multiple computations in a single CPU cycle \cite{zhang2018high}.  

Convolution using GEMM has two major disadvantages: (a) it requires packing overlapping image blocks whose sizes correspond to that of the kernel into the columns of a large temporary matrix. The temporary matrix grows as the number of overlapping image blocks increases. When the kernel and stride are smaller, as is typically the case in Deep Neural Networks (DNNs), there is an increase in the number of image blocks. It results in memory overhead and inefficient memory access, and (b) due to their irregular dimensions, GEMM does not perform as well on convolutional matrices as on matrices derived from classical high-performance computing applications.


\begin{figure*}[t]
\centering
\includegraphics[width=1\linewidth]{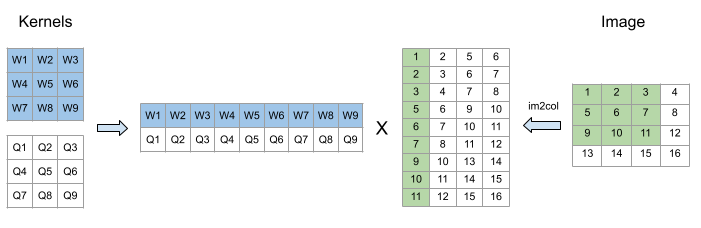}
\caption[Im2col operation example]{Im2col operation (the arrow on the right) with a $3 \times 3$ kernel on a single input channel image. The product is a matrix of 9 rows and 4 columns. The highlighted slice in the image correspond to the highlighted column. There is a significant overlap between each pair of consecutive columns.}
\label{fig:im2col}
\end{figure*}

 
When considering the implementation of convolutions, there are two possible memory layouts - channels last and channels first ~\cite{channel-last-nvidia}. In channels first, the tensor is arranged as NCHW (N batch size, C number of channels, h is height and w is width) in memory while preserving their dimensions. In channels last, the tensor is arranged as NHWC. Channels first is used as the default configuration in various deep learning frameworks~\cite{pytorch_9015, channel-last-pytorch} and many existing pre-trained models are already available in channels first format.


We investigate an alternative to the matrix multiplication based convolution, and demonstrate that it can be highly efficient for CPU-based architecture for channels first memory layout. We propose a scalar-matrix multiplication and zero packing approach that reduces the memory overhead while allowing CPU optimizations for continuous memory layouts.

Memory-efficient Convolution (MEC) have been proposed to reduce the memory overhead while still using matrix-matrix multiplication~\cite{cho2017mec}. The memory overhead in our approach is comparable to that in~\cite{cho2017mec}. By using scalar matrix multiplication without packing, we show that our method can significantly accelerate the computation.



This paper makes the following contributions:
\begin{itemize}
    \item We propose scalar-matrix multiplication with zero packing for convolution, rather than the widely-used approach of matrix-matrix multiplication with column-based packing. 
    \item We show that our approach can accelerate convolution for CPU-based architecture, outperforming im2col+GEMM and state-of-the-art memory efficient convolution (MEC)~\cite{cho2017mec}. This demonstrates that existing models can be executed more efficiently on mobile and low-power devices.
    
\end{itemize}

\section{Background}
\label{sec:formatting}

There are two ways to perform convolutions: (a) by transforming the weights and data into a different space, applying simple operations (such as multiplication), and then transforming back; or (b) by performing direct convolution on the input weights and activation tensors. The FFT and Winograd transforms are examples of the first type. The last type is associated with GEMM-based or high-performance direct convolution implementations. 

\subsection{Notations}

Notations used in this paper are listed in table \ref{tbl:notations}. Consider a convolutional layer which accepts a tensor of $c_i$ channels $\times$ $h$ height $\times$ $w$ width. The layer performs convolution with $c_o$ kernels, each is of $c_i$ $\times$ $k_h$ height $\times$ $k_w$ width. The output is a tensor of $c_o$ channels $\times$ $h'$ height $\times$ $w'$ width. Each output pixel is a linear combination of $c_i*k_h*k_w$ input pixels.

\begin{table}
\centering
\begin{tabular}{ll}
\toprule
$c_i$ & \# Input Channels \\
$c_o$ & \# Output Channels \\
\midrule
$h$   & Input Tensor Height \\
$w$   & Input Tensor Width \\
\midrule
$h'$  & Output Tensor Height\\
$w'$  & Output Tensor Width \\
\midrule
$k_h$ & Kernel Tensor Height \\
$k_w$ & Kernel Tensor Width \\
\midrule
I & Input Tensor \\
K & Kernel Tensor  \\
O & Output Tensor  \\
\midrule
$ T_j^c $ & Sub-matrix of I, $I[c,1:h,j:j+w'-1]$\\
\bottomrule
\end{tabular}
\caption[Notations]{
Notations}
\label{tbl:notations}
\end{table}

\subsection{GEMM-based implementation}

The primary method to compute convolutions without transforms in channels first layout is based on GEMM. GEMMs are a fundamental building block for many operations in neural networks, mainly due to its efficiency. For convolutions, using GEMM performs the same number of math operations as a direct convolution and hence is computationally equivalent.

In order to use GEMM, the tensor is needed to be packed into a matrix. For that, im2col operation~\cite{yanai2016efficient} packs image blocks into columns of a matrix, and the kernel weights are formed into rows of a matrix. ~\Cref{fig:im2col} shows an example. Specifically, each image block (green background) is packed by im2col into a single column. Each kernel (blue background) is a single row. This results in  $(c_o) \times (c_i * k_h * k_w)$ and $(c_i * k_h * k_w) \times (h' * w')$ matrices multiplication.  

The size of the packed image matrix, $(c_i * k_h * k_w) \times (h' * w')$, can be considerably larger than the original image matrix. This is due to the fact that the packed image blocks are overlapping in the original image, resulting in duplication, which incurs a significant memory overhead.

Memory-efficient Convolution (MEC) have been proposed to reduce the memory overhead  by packing multiple columns at once rather than each single individual sub-matrix~\cite{cho2017mec}. We compare our approach with both the aforementioned methods.

\subsection{Transform-based implementation}

The most common transformed used are FFT~\cite{nussbaumer1981fast} and Winograd~\cite{winograd1980arithmetic}.

\begin{itemize}
    \item FFT-based convolution is based on the fact that the Fourier transform of the convolution of two signals is the point-wise multiplication of their Fourier transforms. However, the two signals must be of the same size and therefore the kernels must be padded to the same size as the input tensor. This incurs memory penalty which becomes quite large when kernels are small (e.g., $3 \times 3$), as commonly is the case.
    
    \item The Winograd convolution has been shown to be efficient for small kernels~\cite{lavin2016fast} due to the fact that on modern processors, addition is more efficient than multiplication. The method uses less memory than FFT-based convolution and greatly reduces the number of multiplication operations in convolutions, at the expense of an increase in the number of addition operations. 
    
\end{itemize}

\subsection{Direct Convolution}

A high performance implementation of direct convolution has been proposed \cite{zhang2018high}. They showed that it can outperform a GEMM based convolution in terms of amount of actual performance, parallelism, and reduced memory overhead. However, their method is only applicable for channels last memory layout.

\subsection{Approximated Convolution}

Various approximation for full convolution have been proposed,  including low-rank for efficient computation \cite{Rigamonti_2013_CVPR, jaderberg2014speeding, ben2007gray} and binary neural networks ~\cite{qin2020binary}. In contrast to our approach, the approximation based methods results in degraded accuracy.

\begin{figure*}[t]
\centering
\includegraphics[width=0.75\linewidth]{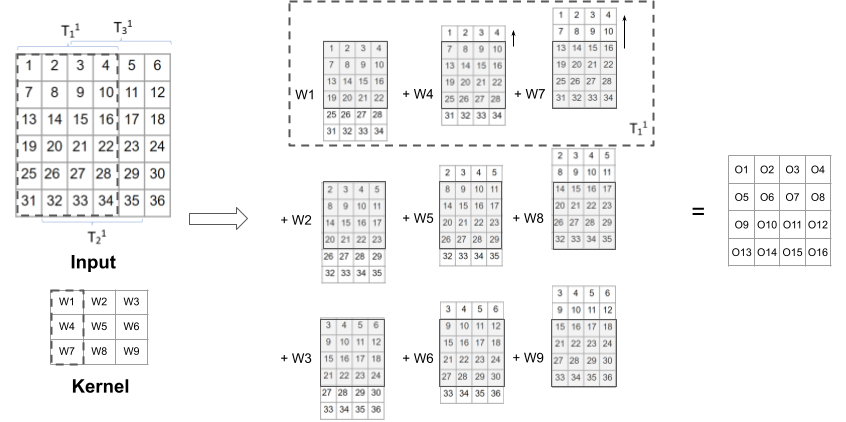}
\caption{Our approach. The result of convolutions of $9$ consecutive positions with a $3 \times 3$ kernel can be viewed as a linear combination of shifted sub-matrices. We extract a sub-matrix of the input tensor and use scalar matrix multiplication with shifted blocks to compute the results.}
\label{fig:our_apporach}
\end{figure*}

\section{SMM-Conv}
\label{Method} 

\subsection{Motivation}

Conventional wisdom suggests that GEMM is well suited for convolution due to the fact that the overhead involved in the preparation phase is well compensated by the highly efficient performance of the matrix multiplication. Existing methods focus on reducing the memory overhead while still applying matrix matrix multiplication for the computation of the convolution. SMM-Conv accelerates the computation of the convolution by addressing both components in the pipeline: it employs scalar matrix multiplication rather than matrix matrix multiplication, and reduces overhead to approximately one copy of the output tensor while reusing the same memory buffer.

\subsection{Our Approach}

In the following, we describe our algorithm with respect to column-major order. Details regarding row-major order derived in a similar manner. 

Given an input tensor I of size $c_i \times h \times w$, and convolutional layer with $c_o$ kernels K each of size $c_i \times k_h \times k_w$, the output tensor O is of size $c_o \times h' \times w'$.\\  

\subsubsection{One input one output channel} The $2D$ output of convolution of an input tensor I of size $h \times w$ with a kernel K of size $k_h \times k_w$ can be considered as summation of \emph{$k_h*k_w$} shifted versions of the input tensor I, with corresponding sub-matrices of size $h' \times w'$ multiplied by corresponding coefficient. Therefore, instead of packing each image block of size $k_h \times k_w$ into a column of size  $(k_h * k_w) \times 1$, we consecutively extract the sub-matrices $T_j^1, j \in [k_w]$ (superscript $c$ is one channel) which consist of all the rows of the I and $w'$ columns, $I[1, 1:h,j:j+w'-1]$ and multiply each sub-matrix of size $h' \times w'$ in $T^1_j$ with the corresponding kernel weight and sum.




~\Cref{fig:our_apporach} present an example for a $3 \times 3$ kernel: the input tensor (image) I is "sliced" to $T^1_1$, $T^1_2$ and $T^1_3$. The $h'\times w'$ sub-matrices of $T^1_j$ (highlighted) are multiplied with corresponding weights of the kernel. A key property of our approach is that we reuse the same memory buffer of size $h \times w'$  to compute the result of the convolution.  The consecutive multiplications with each window within $T^1_j$ access a contiguous region in the memory block of $h' \times w'$ of floating points and not requiring further computation. We call this phase "shifting" as it only requires pointer-arithmetic operations.


\subsubsection{Multiple input and output channels}

We extend the previous algorithm to the multiple channels' case. For that, we loop on the input channels. We consecutively extract  $k_w$ sub-matrices $T_j^c$ which consist of all the rows and $w'$ columns of channel c, $I[c,1:h,j:j+w'-1]$. For each matrix we shift $k_h$ times, obtaining $h'\times w'$ matrices for $k_h * k_w$ scalar-matrix multiplications. For each output kernel $c_o$ we accumulate the result into the corresponding output channel. This is done repeatedly for all the $c_i$ input channels.

For contiguous access, we use a kernel layout of $c_i \times k_w \times k_h \times c_o$ multidimensional array. Notice that the ordering of dimensions adapted to match the order of the access of the algorithm.

The algorithm is shown in~\cref{alg:fastconv2dalgo}.

\subsubsection{Single thread vs. Parallel.}

Our convolution implementation is divided to two steps: extracting the input tensor into $T_j^c$ sub-matrices and scalar-matrix multiplications.

For fast parallel algorithms, we adhere to the following principles:
\begin{itemize}
    \item \textbf{Memory invalidation.} Writing and reading from a memory block simultaneously is discouraged as it could result in invalid readings and prevents  CPU caching.
    \item \textbf{Parallel writing.} All output elements should be computable in parallel.
\end{itemize}

For $d$ threads ($1 <= d$), $d$ memory buffers are allocated. Each memory buffer is a $h \times w'$ matrix. Each thread is associated with a single memory buffer and $c_o / d$ output feature maps.

We iterate for $c_i * k_w / d$ times and associate every memory buffer with an input channel $c$ and horizontal offset $j$ ($1 <= c <= c_i, 1 <= j <= k_w)$. Each thread extract $T_j^c$ into its associated memory buffer. Then each thread performs scalar-matrix multiplications with every $h'\times w'$ shifted window of each $T_j^c$ computed before into the thread's associated $c_o / d$ output feature maps. The algorithm is shown in~\cref{alg:threadedpacking}.





\subsection{Memory Requirements}

In our implementation, $T^i_j$ are written into the same memory buffer. After extraction, scalar matrix multiplications are executed on every $h' \times w'$ slice of the matrix and accumulated into $c_o$ output matrices.

Im2col routine, on the other hand, packs every $h' \times w'$ slice of $I$ and requires a $c_i * k_h * k_w \times h' \times w'$ tensor for its output.

Comparing the ratio between the memory required for our implementation and for im2col:
\begin{equation}
    \frac{c_i*k_h*k_w*h'*w'}{h * w'} = c_i*k_h*k_w \frac{h'}{h}   
\end{equation}

In many commonly used convolutional layers such as~\cite{redmon2016you, huang2018yolo}, $ h' \approx h $. In conclusion, im2col requires approximately $c_i*k_h*k_w$ times the memory used by our algorithm.

\subsection{Implementation advantages}
\label{sec:ia}
SMM-Conv extract a sub-matrix, iteratively preforms shifting operation, scalar matrix multiplication and summation. Usage of a contiguous memory buffer for short steps rather than matrix multiplication subroutine is beneficial for the following assumed reasons:
\begin{itemize}
    \item \textbf{\emph{FMA} instructions.} Performing $h'*w'$ multiplications and accumulations with a contiguous floating points memory buffer benefits from the fused multiplication-accumulation \emph{SIMD} operation \cite{zhang2018high}.
    
    \item \textbf{Memory demand.} Available memory resources for low-power embedded devices are expensive. SMM-Conv reduces the total temporary memory by $c_i * K_h * k_w$. 
    
    \item \textbf{CPU caching} During the entire execution span, we store only a matrix of $h*w'$ and use it exclusively for reading, without loading and unloading. This type of configuration is well suited for caching. 
\end{itemize}

\begin{algorithm}[t]
\caption{Single-threaded SMM convolution}
\label{alg:fastconv2dalgo}
\SetAlgoLined
\SetNoFillComment
\KwIn{$I$ - a $c_i \times h \times w$ input tensor}
\myinput{$K$ - a $c_i \times k_w \times k_h \times c_o$ kernel tensor}
\KwResult{$O$ - a $c_o \times h' \times w'$ output tensor}
Set O values to zero\\
\For{ $c \gets 1$ to $c_i$ }{

    \For{ $j \gets 1$ to $k_w$ } {
        
        $Sliced\_Mat \gets T_j^c$ \\
        \ \\
        
        \For{ $k \gets 1$ to $k_h$ } {
            \tcc{Shifting}
            $Shifted\_Mat \gets Sliced\_Mat[k:h'+k, :] $ \\
            \ \\
          
            \tcc{Scalar-Matrix multiplication and accumulation}
          \For{ $m \gets 1$ to $c_o$ } {
            $w \gets K[c, j, k, m] $\\
            $O[c, :, :] += w * Shifted\_Mat $
          }
        }
    }
}
\end{algorithm}

\begin{algorithm}[t]
\caption{Parallel SMM convolution}
\label{alg:threadedpacking}
\SetAlgoLined
\SetNoFillComment
\SetKw{KwTL}{Thread Limit}
\SetKw{KwTN}{Thread numbering}
\KwIn{$I$ - a $c_i \times h \times w$ input tensor}
\myinput{$K$ - a $c_i \times k_w \times k_h \times c_o$ kernel tensor}
\KwResult{$O$ - a $c_o \times h' \times w'$ output tensor}
\KwTL{using $d$ threads.}\\
\KwTN{\#n := current thread number}

  set O values to zero.\\
  \For{ $\ell \gets 1$ to $c_i * k_w / d$ } {
    \tcc{Associate input channel and horizontal offset to buffer \#n}
    $Sliced\_mat\_channel^{\#n} \gets $ input channel c \\
    $Sliced\_mat\_offset^{\#n} \gets j$ \\
    
    \tcc{Parallel packing into a $h \times w'$ matrix}
    $Sliced\_Mat^{\#n} \gets T_j^c$ \\
      
      \ \\
      \textit{thread-sync} \\
      \ \\
      
      \For { $\mu \gets 0$ to $d - 1$ } {
          \For{ $k \gets 1$ to $k_h$ } {
            \tcc{Shifting}
            $Shifted\_Mat^{\mu} \gets Sliced\_Mat^{\mu}[k:h'+k, :] $ \\
            \ \\
            \tcc{Scalar-Matrix multiplication and accumulation. 
            Each thread writes to $c_o / d$ output channels}
            \For{ $\lambda \gets 1$ to $c_o / d$ }{
            
              $w \gets K[c, j, k, \lambda * \#n] $\\
              $O[\lambda * \#n, :, :] += 
                w * Shifted\_Mat^{\mu} $ 
            }
          }
      }
    
  }

\end{algorithm}

\section{Experimental Results}
\label{Experiments} 

In this section, we present performance results of our SMM-conv convolution implementation against existing convolution approaches.

\subsection{Experimental Setup}

\textbf{Baselines} We compare SMM-Conv with im2col+GEMM and MEC~\cite{cho2017mec}. We implemented our approach in C++ using OpenMP \cite{openmp08}. For CPU multi-threaded application, we use im2col+GEMM implemented by PyTorch \cite{pytorch_9015} C++ API, which uses the Intel’s Math Kernel Library (MKL) \cite{intelmkl}.  For embedded devices and single thread application, we implemented direct convolution, im2col and \emph{GEMM} based on the PyTorch implementation. For MEC~\cite{cho2017mec}, we used their available code.  We ran our experiments on Intel Core i7-1165G7 CPU with 4 cores and 8 logical processors.

\subsection{Performance}
\begin{figure*}[t]
\centering
\includegraphics[
    width=1.0\linewidth,
    height=0.49\linewidth
    ]{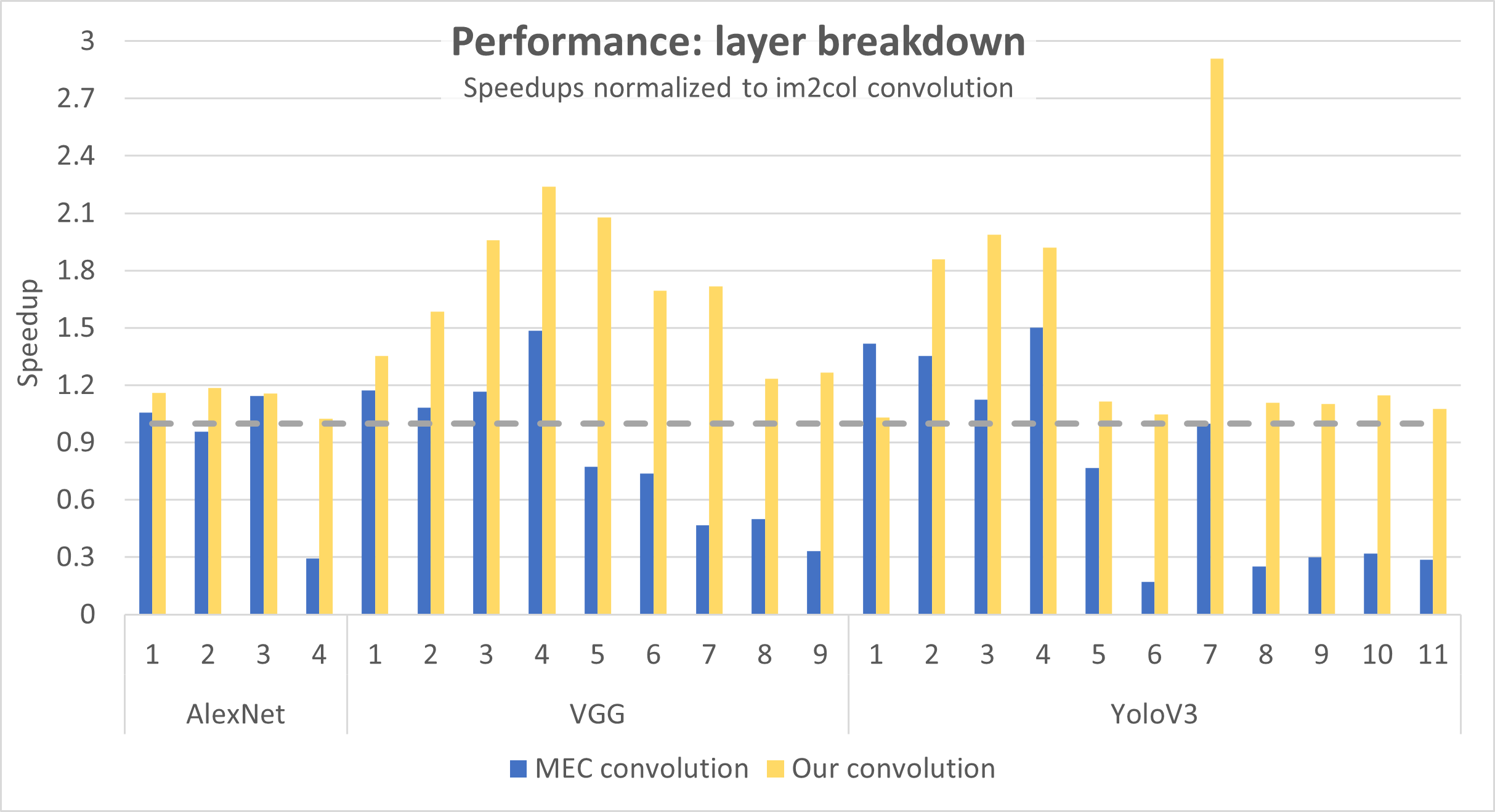}
\caption[Speedup to various convolution neural networks]{Acceleration of convolutional layers in various neural networks. The x-axis is the depth of the layer and the y-axis is the speedup, normalized to im2col convolution.}
\label{fig:networks}
\end{figure*}

All implementations were ran against all convolutional layers found in AlexNet \cite{krizhevsky2012imagenet},
VGG \cite{Simonyan15} and YoloV3 \cite{huang2018yolo}.  The different convolutional layers in these three CNNs span a wide range of sizes of input, output and kernel weights. They are also commonly used as benchmarks for demonstrating the performance of convolution implementations. Overall, our convolution outperforms both im2col-based convolution and MEC. See~\Cref{tbl:networks} for execution time of ours against im2col convolution and MEC, on whole network execution duration. ~\Cref{fig:networks} presents the layer breakdown with respect to the baselines. The relative performance of the different implementations is normalized to the im2col convolution (incl.  \emph{GEMM} routine). It can be seen that SMM-Conv can gain a speedup of up to 200\% with respect to a specific layer.  The different methods share a similar amount of multiplications and accumulations. The speedup of SMM-Conv is due to its efficient use of scalar matrix multiplication (See~\cref{sec:ia}).

\begin{table}
\centering
\begin{tabular}{l l l l l}
\toprule
\textbf{Network} & \textbf{Im2col} & \textbf{MEC} & \textbf{Ours} & \textbf{Speedup} \\
\midrule
AlexNet & 0.4608 & 0.2008 &\textbf{0.1348} & 3.4183 \\
VGG     & 2.3670 & 2.8562 & \textbf{1.3535} & 2.1102 \\
YoloV3 &  0.4478 & 0.5779 & \textbf{0.2889} & 2.0003 \\
\bottomrule\\
\end{tabular}
\caption[Speedup to various convolution neural networks]{
Various convolution neural networks' execution times and speedups (in seconds).}
\label{tbl:networks}
\end{table}

\subsection{Model Scalability}
We compare SMM-Conv  to im2col and MEC with different convolutional layer parameters. The relative performance is normalized to the \emph{GEMM} routine + im2col packing method.

\subsubsection{Input channels count}
\begin{figure*}[t]
\centering
\includegraphics[
    width=1.0\linewidth,
    height=0.48\linewidth
    ]{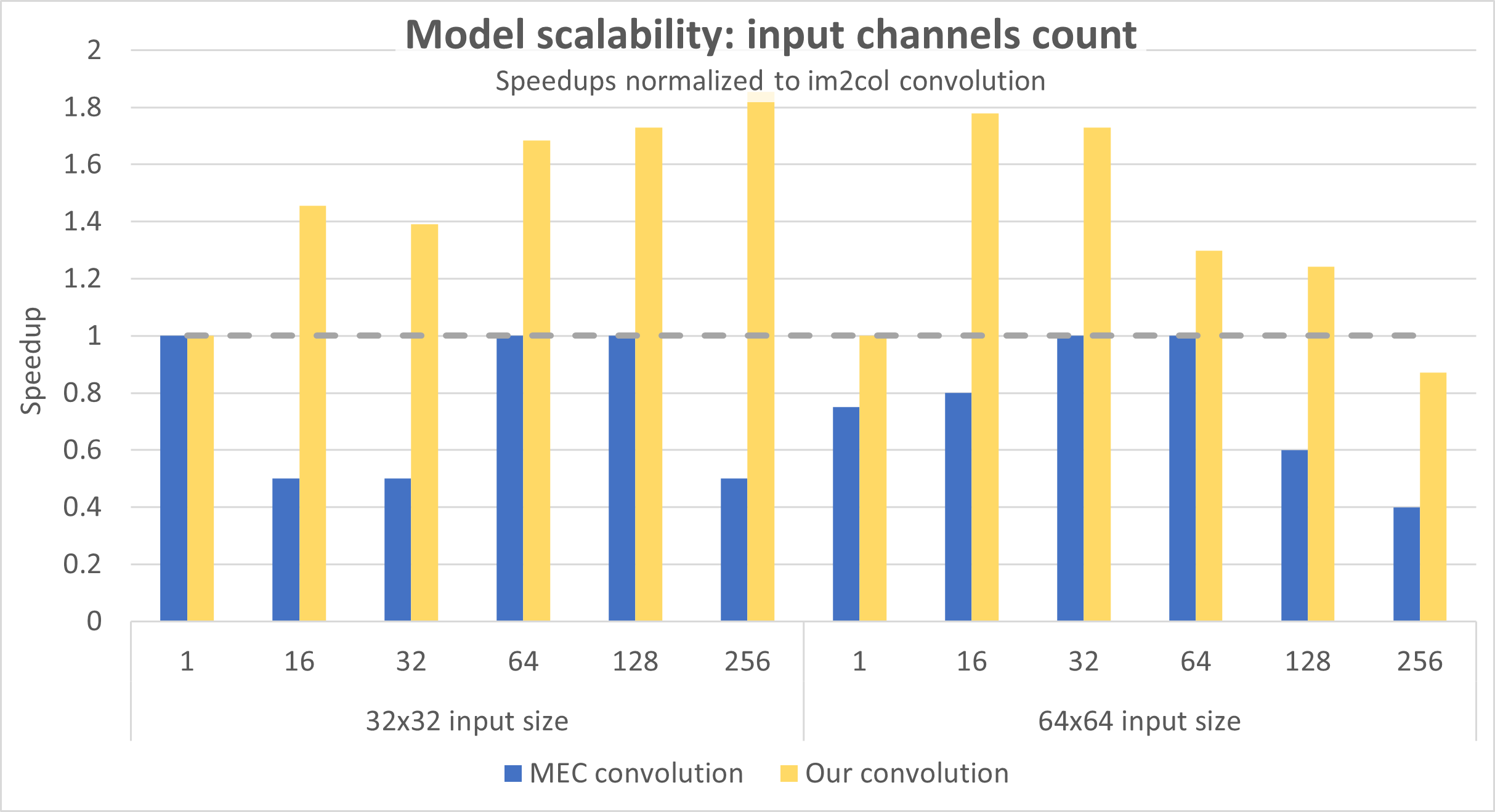}
\caption[Speedups of various number of input channels]{
Acceleration of input channels. The x-axis is the number of input channels and the y-axis is the speedup, normalized to im2col convolution. }
\label{fig:in_channels}
\end{figure*}

In this experiment, we compared 1, 16, 32, 64, 128 and 256 input channels on $32\times 32$ and $64\times 64$ input dimensions, $3 \times 3$ kernels, and 32 output channels. See figure \ref{fig:in_channels}.

While SMM-Conv memory is indifferent to the amount of input channels, im2col convolution and MEC require a memory block that is affected by the amount of input channels.

\subsubsection{Input spatial dimensions}

\begin{figure*}[t]
\centering
\includegraphics[
    width=1.0\linewidth,
    height=0.48\linewidth
    ]{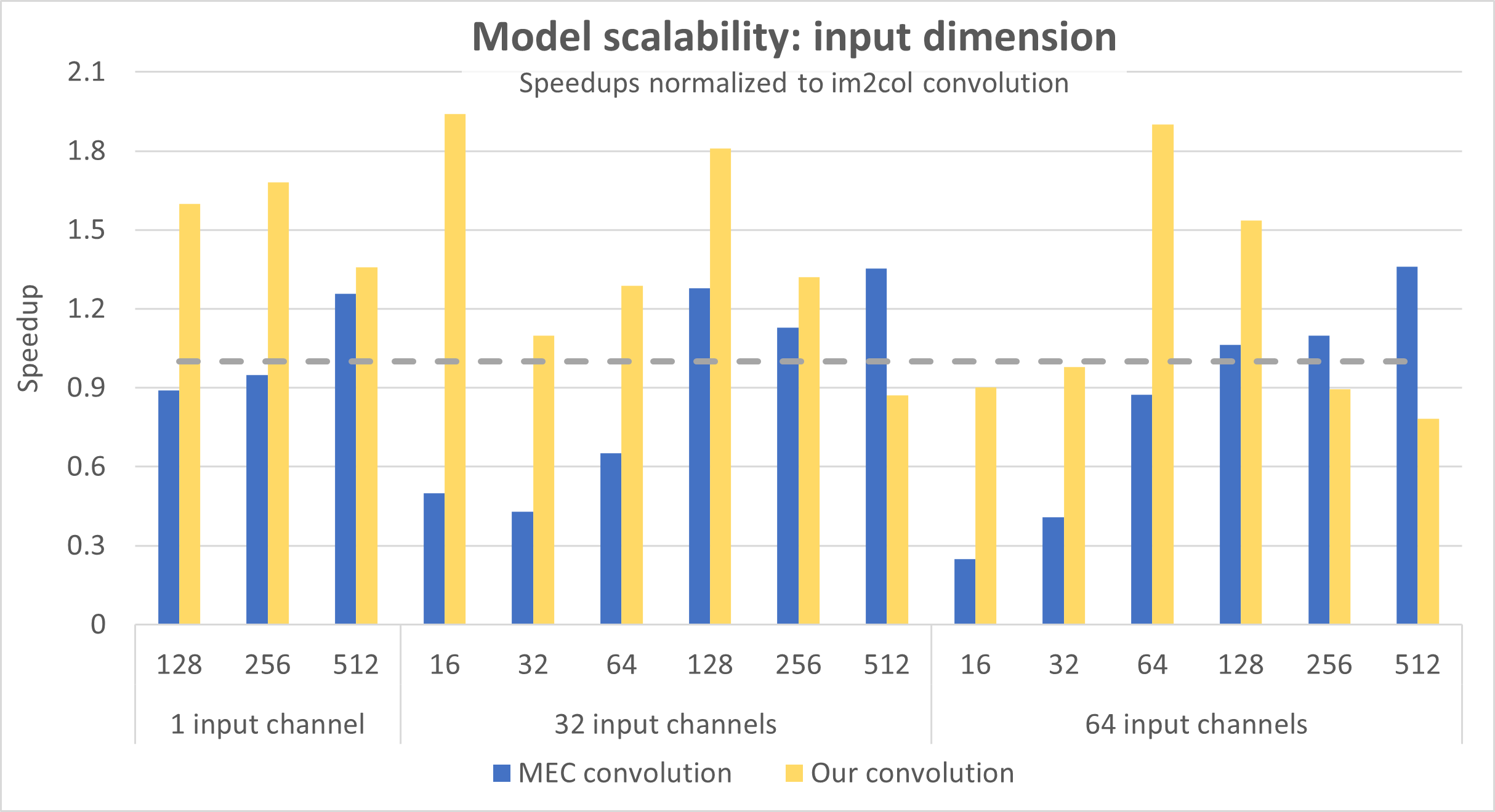}
\caption[Speedups of various number of input dimensions]{
A comparison of the speedups of different squared input dimensions. The x-axis represents the first dimension of the input, and the y-axis represents the speedup, normalized to im2col convolution.}
\label{fig:input_dim}
\end{figure*}

In this experiment we compared $ 32 \times 32, 64 \times 64, 128 \times 128, 256 \times 256$ and $512 \times 512$ input dimensions. We compared 1, 32 and 64 input channels, 32 output channels, $3 \times 3$ kernels. 
See figure \ref{fig:input_dim}.  

The runtime of im2col packing is negligible as large memory copying throughput is high (by using techniques such as streaming) and the majority of the execution time is spent on multiplication. 
SMM-Conv number of matrix extractions, shiftings and scalar-matrix multiplications is determined by the kernel size while MEC's number of required packings in each steps is determined by $H'$ and $W'$.


\subsubsection{Kernel sizes}

\begin{figure*}[t]
\centering
\includegraphics[
    width=1.0\linewidth,
    height=0.48\linewidth
    ]{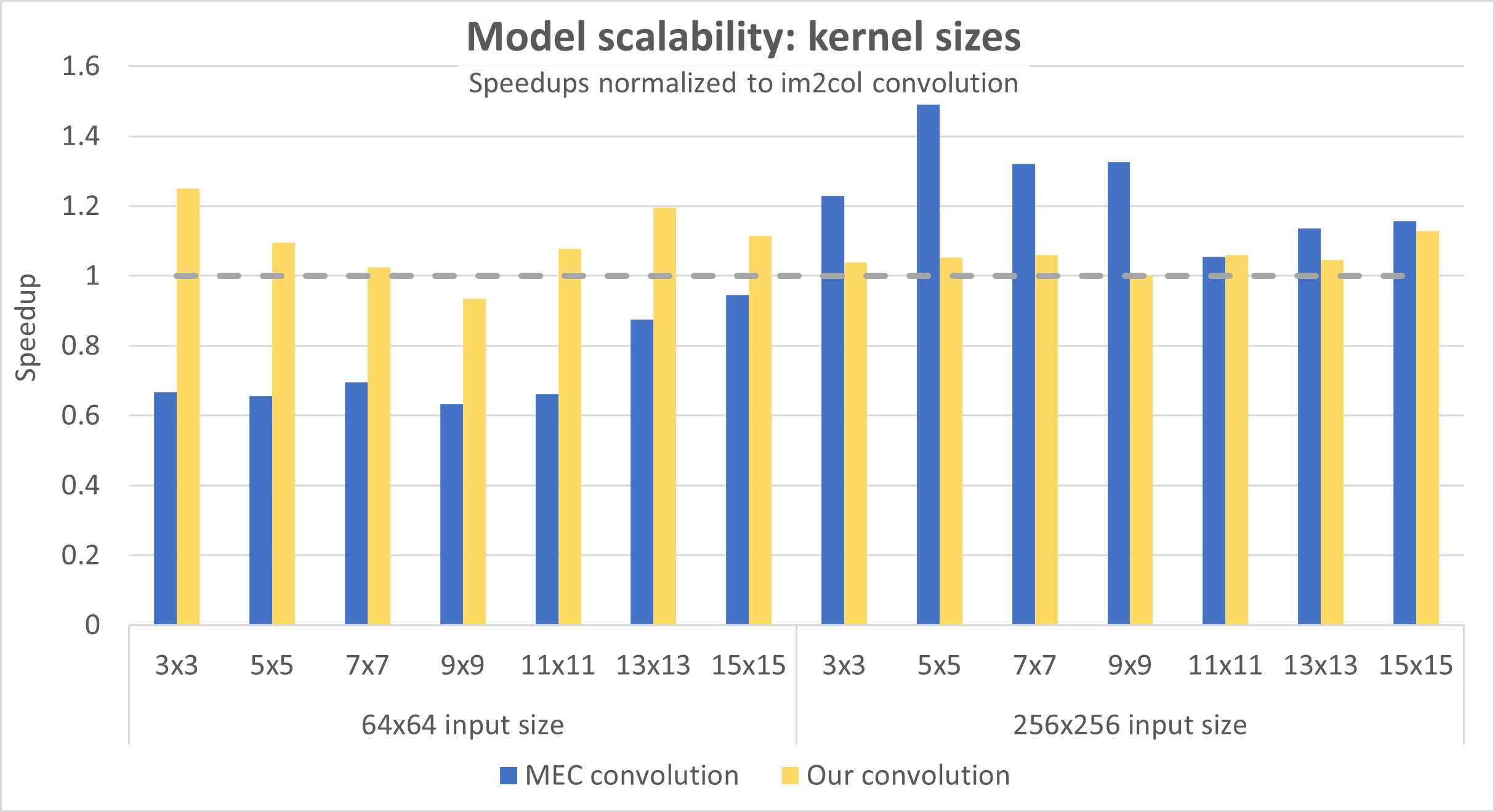}
\caption[Speedups of various number of kernel sizes]{Speedups of various kernel sizes. The x-axis represents the size of the kernels, and the y-axis represents the speedup, normalized to im2col convolution.}
\label{fig:kernel_sizes}
\end{figure*}

In this experiment we compared  $3 \times 3, 5 \times 5, 7 \times 7, 9 \times 9, 11 \times 11, 13 \times 13$ and $15 \times 15$ kernels on $64 \times 64$ and $256 \times 256$ input sizes, 32 input channels and 32 output channels. See figure \ref{fig:kernel_sizes}.

Im2col output matrix has  $c_i * k_w * k_h$ rows and $w'*h'$ columns and therefore grows as the kernel size increased. SMM-Conv memory block, of length $h*w'$, get smaller as the kernel expands in the horizontal direction and is indifferent to kernel height changes. 
MEC memory block has $h'$ rows and $h*k_w$ columns, and therefore if $h >> k_h$ the memory block grows as the kernel expands.


\subsubsection{Output channels count}
\begin{figure*}
\centering
\includegraphics[
    width=1.0\linewidth,
    height=0.49\linewidth
    ]{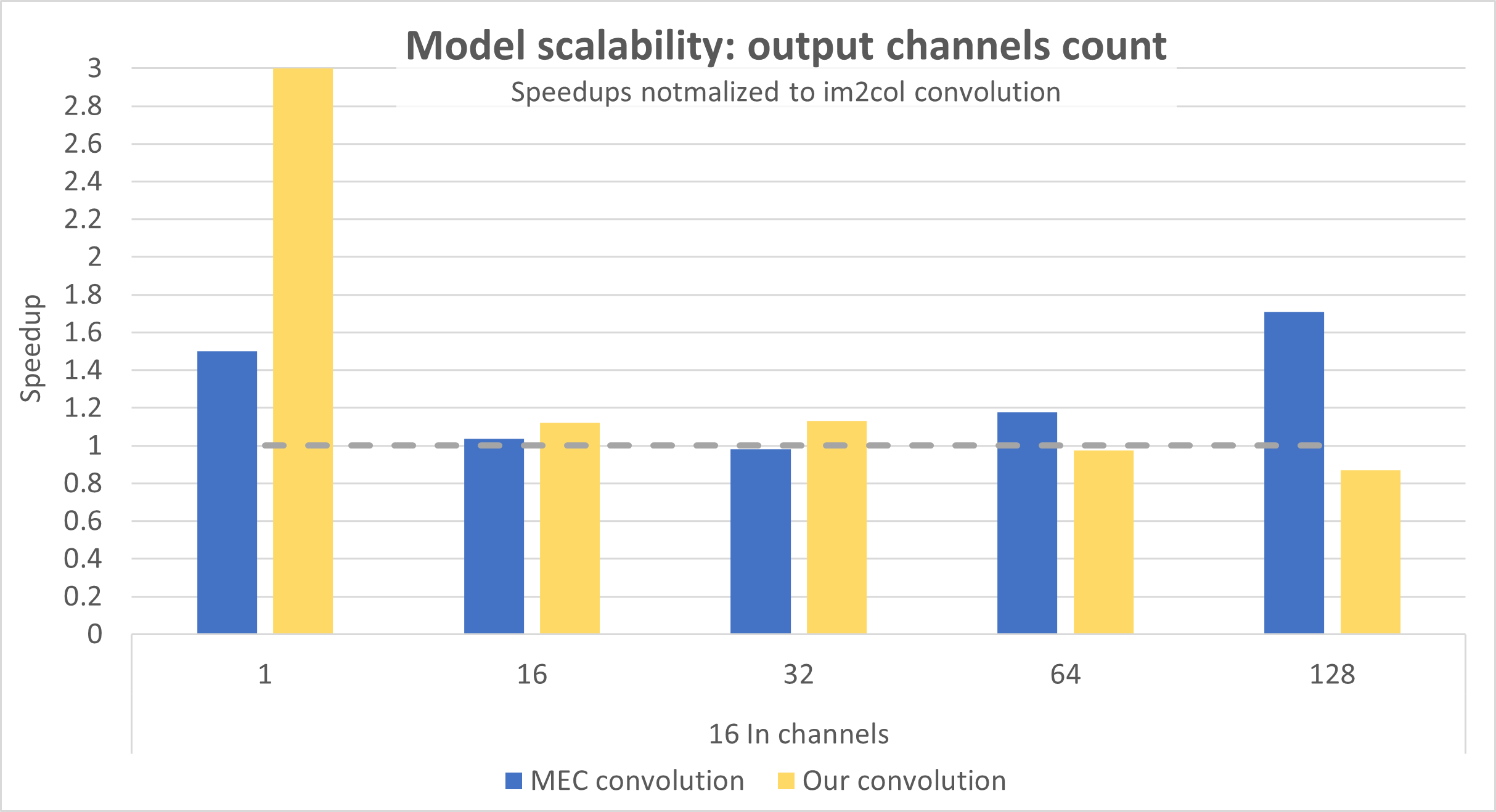}
\caption[Speedups of various number of output channels]{
Speedups of various number of output channels.  The x-axis represents the number of output channels, and the y-axis represents the speedup, normalized to im2col convolution.}
\label{fig:output_channels}
\end{figure*}

In this experiment we compared 1, 8, 16, 32, 64 and 128 output channels on $256 \times 256$ input dimension, $3 \times 3$ kernels and 16 input channels. See figure \ref{fig:output_channels}. The speedup of SMM-Conv shown in figure \ref{fig:output_channels} for single output channel is due to our zero packing, which is negligible for increased number of output channels.

\section{Conclusion}

We presented SMM-Conv for faster convolution for embedded and low-powered devices. Our approach, unlike existing methods, is based on scalar matrix multiplication and does not require packing at all. We showed that SMM-Conv can accelerate convolution for commonly used architectures, including YOLO, AlexNet and VGG.  SMM-Conv can be easily implemented, allowing deployment for various existing deep learning frameworks and existing pre-trained models.

{\small
\bibliographystyle{ieee_fullname}
\bibliography{egbib}

\begin{thebibliography}{10}\itemsep=-1pt

\bibitem{10095537}
Shai Aharon and Gil Ben-Artzi.
\newblock Hypernetwork-based adaptive image restoration.
\newblock In {\em ICASSP 2023 - 2023 IEEE International Conference on
  Acoustics, Speech and Signal Processing (ICASSP)}, pages 1--5, 2023.

\bibitem{ben2007gray}
Gil Ben-Artzi, Hagit Hel-Or, and Yacov Hel-Or.
\newblock The gray-code filter kernels.
\newblock {\em IEEE transactions on pattern analysis and machine intelligence},
  29(3):382--393, 2007.

\bibitem{blackford2002updated}
L~Susan Blackford, Antoine Petitet, Roldan Pozo, Karin Remington, R~Clint
  Whaley, James Demmel, Jack Dongarra, Iain Duff, Sven Hammarling, Greg Henry,
  et~al.
\newblock An updated set of basic linear algebra subprograms (blas).
\newblock {\em ACM Transactions on Mathematical Software}, 28(2):135--151,
  2002.

\bibitem{cho2017mec}
Minsik Cho and Daniel Brand.
\newblock Mec: memory-efficient convolution for deep neural network.
\newblock In {\em International Conference on Machine Learning}, pages
  815--824. PMLR, 2017.

\bibitem{9380520}
Yacov Hel-Or and Gil Ben-Artzi.
\newblock The role of redundant bases and shrinkage functions in image
  denoising.
\newblock {\em IEEE Transactions on Image Processing}, 30:3778--3792, 2021.

\bibitem{huang2018yolo}
Rachel Huang, Jonathan Pedoeem, and Cuixian Chen.
\newblock Yolo-lite: a real-time object detection algorithm optimized for
  non-gpu computers.
\newblock In {\em 2018 IEEE International Conference on Big Data (Big Data)},
  pages 2503--2510. IEEE, 2018.

\bibitem{intelmkl}
Intel.
\newblock Math kernel library https://software.intel.com/en-us/intel-mkl, 2015.

\bibitem{jaderberg2014speeding}
Max Jaderberg, Andrea Vedaldi, and Andrew Zisserman.
\newblock Speeding up convolutional neural networks with low rank expansions.
\newblock {\em arXiv preprint arXiv:1405.3866}, 2014.

\bibitem{jia2014caffe}
Yangqing Jia, Evan Shelhamer, Jeff Donahue, Sergey Karayev, Jonathan Long, Ross
  Girshick, Sergio Guadarrama, and Trevor Darrell.
\newblock Caffe: Convolutional architecture for fast feature embedding.
\newblock In {\em Proceedings of the 22nd ACM international conference on
  Multimedia}, pages 675--678, 2014.

\bibitem{krizhevsky2012imagenet}
Alex Krizhevsky, Ilya Sutskever, and Geoffrey~E Hinton.
\newblock Imagenet classification with deep convolutional neural networks.
\newblock In {\em Advances in neural information processing systems}, pages
  1097--1105, 2012.

\bibitem{lavin2016fast}
Andrew Lavin and Scott Gray.
\newblock Fast algorithms for convolutional neural networks.
\newblock In {\em Proceedings of the IEEE conference on computer vision and
  pattern recognition}, pages 4013--4021, 2016.

\bibitem{nussbaumer1981fast}
Henri~J Nussbaumer.
\newblock The fast fourier transform.
\newblock In {\em Fast Fourier Transform and Convolution Algorithms}, pages
  80--111. Springer, 1981.

\bibitem{channel-last-nvidia}
Nvidia: Deep learning performance documentation; tensor layouts.
\newblock
  \url{https://docs.nvidia.com/deeplearning/performance/dl-performance-convolutional/index.html}.

\bibitem{openmp08}
{OpenMP Architecture Review Board}.
\newblock {OpenMP} application program interface version 3.0, May 2008.

\bibitem{pytorch_9015}
Adam Paszke, Sam Gross, Francisco Massa, Adam Lerer, James Bradbury, Gregory
  Chanan, Trevor Killeen, Zeming Lin, Natalia Gimelshein, Luca Antiga, Alban
  Desmaison, Andreas Kopf, Edward Yang, Zachary DeVito, Martin Raison, Alykhan
  Tejani, Sasank Chilamkurthy, Benoit Steiner, Lu Fang, Junjie Bai, and Soumith
  Chintala.
\newblock Pytorch: An imperative style, high-performance deep learning library.
\newblock In H. Wallach, H. Larochelle, A. Beygelzimer, F. d\textquotesingle
  Alch\'{e}-Buc, E. Fox, and R. Garnett, editors, {\em Advances in Neural
  Information Processing Systems 32}, pages 8024--8035. Curran Associates,
  Inc., 2019.

\bibitem{channel-last-pytorch}
Pytorch: Channels last memory format.
\newblock
  \url{https://pytorch.org/tutorials/intermediate/memory_format_tutorial.html}.

\bibitem{qin2020binary}
Haotong Qin, Ruihao Gong, Xianglong Liu, Xiao Bai, Jingkuan Song, and Nicu
  Sebe.
\newblock Binary neural networks: A survey.
\newblock {\em Pattern Recognition}, 105:107281, 2020.

\bibitem{redmon2016you}
Joseph Redmon, Santosh Divvala, Ross Girshick, and Ali Farhadi.
\newblock You only look once: Unified, real-time object detection.
\newblock In {\em Proceedings of the IEEE conference on computer vision and
  pattern recognition}, pages 779--788, 2016.

\bibitem{Rigamonti_2013_CVPR}
Roberto Rigamonti, Amos Sironi, Vincent Lepetit, and Pascal Fua.
\newblock Learning separable filters.
\newblock In {\em Proceedings of the IEEE Conference on Computer Vision and
  Pattern Recognition (CVPR)}, June 2013.

\bibitem{Simonyan15}
Karen Simonyan and Andrew Zisserman.
\newblock Very deep convolutional networks for large-scale image recognition.
\newblock In {\em International Conference on Learning Representations}, 2015.

\bibitem{vanhouckeimproving}
V Vanhoucke and MZ Mao.
\newblock Improving the speed of neural networks on cpus.[(accessed on 1 may
  2019)]; deep learning \& unsupervised feature learning workshop nips.

\bibitem{winograd1980arithmetic}
Shmuel Winograd.
\newblock {\em Arithmetic complexity of computations}, volume~33.
\newblock Siam, 1980.

\bibitem{yanai2016efficient}
Keiji Yanai, Ryosuke Tanno, and Koichi Okamoto.
\newblock Efficient mobile implementation of a cnn-based object recognition
  system.
\newblock In {\em Proceedings of the 24th ACM international conference on
  Multimedia}, pages 362--366, 2016.

\bibitem{zhang2018high}
Jiyuan Zhang, Franz Franchetti, and Tze~Meng Low.
\newblock High performance zero-memory overhead direct convolutions.
\newblock In {\em International Conference on Machine Learning}, pages
  5776--5785. PMLR, 2018.

\end{thebibliography}
}

\end{document}